\documentclass[runningheads, envcountsame, a4paper]{llncs}

\usepackage{amsmath}
\usepackage{cite}
\usepackage{multirow}
\usepackage[caption=false]{subfig}

\usepackage[misc]{ifsym}
\usepackage{algorithm2e}
\usepackage[pdftex]{graphicx}
\usepackage{epstopdf}
\usepackage[hyphens]{url}
\usepackage{microtype}

\begin{document}

\title{P2ExNet: Patch-based Prototype Explanation Network}

\author{Dominique Mercier\inst{,1,2}\Letter \and
Andreas Dengel \inst{1,2} \and
Sheraz Ahmed \inst{2}}

\institute{Computer Science Department, Technische Universität Kaiserslautern,\\
67663 Kaiserslautern, Germany \and
Smart Data and Knowledge Services, German Research Center for Artificial Intelligence (DFKI), 67663 Kaiserslautern, Germany\\
\email{\{dominique.mercier, andreas.dengel, sheraz.ahmed\}@dfki.de}}

\titlerunning{P2ExNet}
\authorrunning{D. Mercier et al.}

\toctitle{P2ExNet: Patch-based Prototype Explanation Network}
\tocauthor{D. Mercier et al.}

\maketitle

\setcounter{footnote}{0}

\begin{abstract}
Deep learning methods have shown great success in several domains as they process a large amount of data efficiently, capable of solving difficult classification, forecast, segmentation, and other tasks. However, these networks suffer from their inexplicability that limits their applicability and trustworthiness. Although there exists work addressing this perspective, most of the existing approaches are limited to the image modality due to the intuitive and prominent concepts. Unfortunately, the patterns in the time-series domain are more complex and non-comprehensive, and an explanation for the network decision is pivotal in critical areas like medical, financial, or industry. Addressing the need for an explainable approach, we propose a novel interpretable network scheme, designed to inherently use an explicable reasoning process inspired by the human cognition without the need of additional post-hoc explainability methods. Therefore, the approach uses class-specific patches as they cover local patterns, relevant to the classification, to reveal similarities with samples of the same class. Besides, we introduce a novel loss concerning interpretability and accuracy that constraints P2ExNet to provide viable explanations of the data that include relevant patches, their position, class similarities, and comparison methods without compromising performance. An analysis of the results on eight publicly available time-series datasets reveals that P2ExNet reaches similar performance when compared to its counterparts while inherently providing understandable and traceable decisions.

\end{abstract}

\section{Introduction}
Nowadays, deep neural networks are popular and used in many different domains comprising image processing, natural language processing, and time-series processing. Though these deep networks have achieved high performance, they are still black boxes in nature. This behavior makes it tough to understand the reasons behind the decisions. In particular, this black box nature hinders the use of these models in critical domains like medical, autonomous driving, industrial, financial, and raises the need for interpretability methods to provide intuitive and understandable explanations. Only explainable models can are usable in critical domains that require transparency~\cite{samek2017explainable}. 

The existing methods for interpreting decisions of deep learning models are mostly applicable to image modalities. In particular, image concepts are intuitive by default~\cite{zhang2018visual}. Besides the image domain, there is only a limited amount of work in the field of time-series as the modalities are more complex and usually not directly interpretable for a human. Nevertheless, these time-series analysis networks and their explanations are pivotal for their industrial and financial use. Therefore, we propose P2ExNet as an approach to deal with time-series data.

Also, existing approaches are mostly post-hoc methods that are applied after the classifier to explain their decisions~\cite{choo2018visual}. Intuitively, these approaches keep the network as it is without any change to the structure, enabling their use on almost every architecture. Usually, this results in an instance-based local explanation that does not explain any global behavior. In contrast to post-hoc methods, the intrinsic methods focus on model design concerning the inference process to provide an understandable global explanation. Ultimately, neither of the two approaches is superior as both have to deal with several limitations regarding the quality, subjectivity~\cite{lipton2016mythos}, the audience, and the domain usage.

To overcome these limitations, we propose a network architecture for time-series analysis based on the standard deep neural network architecture providing a global explanation using representative class-specific prototypes and an instance-based local explanation using patch-based similarities and class-similarities. The inference process of our architecture follows the human-related reasoning process~\cite{guidoni1985natural} and uses concepts and prototypes~\cite{li2018deep}. Intuitive class-specific patches explain the network decision. Our approach is superior compared to existing template matching approaches~\cite{brunelli2009template} in the manner of generalization and applicability. Our experiments emphasize the use of our network structure by highlighting the comparable performance when compared to a non-interpretable network of the same size over eight publicly available datasets while preserving an intuitive and traceable explanation.

\section{Related Work}
The field of network interpretability covers post-hoc and intrinsic methods. Based on the use-case, it is not always possible to use both methods as these methods come with restrictions concerning the data and the network. In the following paragraphs, we address the perspectives, their advantages, and drawbacks.

\subsection{Post-hoc}
Using post-hoc methods to explain the decisions of deep neural networks is a very prominent approach as these methods do not modify the network architecture and can provide an instance base explanation. Furthermore, these methods offer instance-based as well as global explanations resulting in broad applicability.

\subsubsection{Instance-based:} 
A widespread instance-based post-hoc class of approaches in the field of image domain are so-called back-propagation methods~\cite{bojarski2016visualbackprop}. These approaches produce heat-maps highlighting the most relevant and sensitive parts concerning the network decision. There exist enhancements that evolved~\cite{zintgraf2017visualizing} and take various aspects into account to improve the expressiveness and consistency. Another post-hoc instance-based class of methods are the layer-wise relevance propagation methods~\cite{gu2018understanding, arras2017explaining} that produce results that are close to the heat-maps but more stable. In particular, the image domain explored different approaches to visualize the activations~\cite{yosinski2015understanding} or make use of the gradients~\cite{selvaraju2017grad} or saliency~\cite{simonyan2013deep} to produce heat-maps for instances. However, in the case of the time-series modalities, there exists only a limited amount of work~\cite{siddiqui2019tsviz}.

\subsubsection{Global:} In contrast to instance-based methods, there exist attempts to compute a global behavior based on the influence of the samples~\cite{koh2017understanding, yeh2018representer}. These methods provide an idea of helpful and harmful dataset samples to detect outliers and debug dataset using the sample influence. Another approach is to attach an interpretable architecture to the trained network. As presented in~\cite{Palacio_2018_CVPR}, the attachment of an autoencoder before the neural network and a customized loss function for the autoencoder can enhance the interpretability. Siddiqui et al.~\cite{siddiqui2020tsinsight} presented an adoption of this approach for the time-series domain with an adjusted loss function.

\subsection{Intrinsic}
Intrinsic methods approach the problem from a different perspective by incorporating the interpretability directly. Therefore, they modify the model architecture by introducing interpretable layers~\cite{zhang2018interpretable}. A drawback of these approaches is the restricted learning process that can harm the performance. An intuitive interpretable layer solution are prototype layers to explain model decision~\cite{angelov2019towards}. Mainly, two types of prototypes showed to provide reasonable explanations. First, class prototypes that cover the complete input~\cite{li2018deep, gee2019explaining} and second patch prototypes~\cite{chen2018looks}. 

\subsection{Limitations of Existing Methods}
Even though there exists work to explain the network decisions, most of the approaches are limited to image modalities~\cite{schlegel2019towards}. Furthermore, there is ongoing research investigating the consistency, expressiveness, and subjectivity of these explanations. Some findings prove the inconsistency of saliency-based methods~\cite{tomsett2019sanity} and the expressiveness~\cite{alvarez2018robustness}. Also, methods that use sparsity constraints suffer from the same problems concerning their consistency.

\section{P2ExNet: The Proposed Approach.}
This section provides insights into the proposed approach. It starts with a motivation followed by the general architecture structure, the mathematical background, and the training procedure.

\subsection{Motivation: An Understandable Reasoning Behavior.}
Inspired by human reasoning behavior, we aligned our framework to rely on implicit knowledge about objects and examples already seen before. This approach is similar to the humans' inference process. Precisely, we compare new instances to abstract concepts include class-specific features. The term prototypical knowledge describes the knowledge about these concepts and covers the analogical process to map new to the existing knowledge~\cite{gentner2010analogical}. Following this process, the proposed method uses shallow representations. These prototypes encode class-specific pattern and provide the decision based on similarity. 

\subsection{Architecture}
\begin{figure}[!t]
\centering
\includegraphics[width=0.8\linewidth]{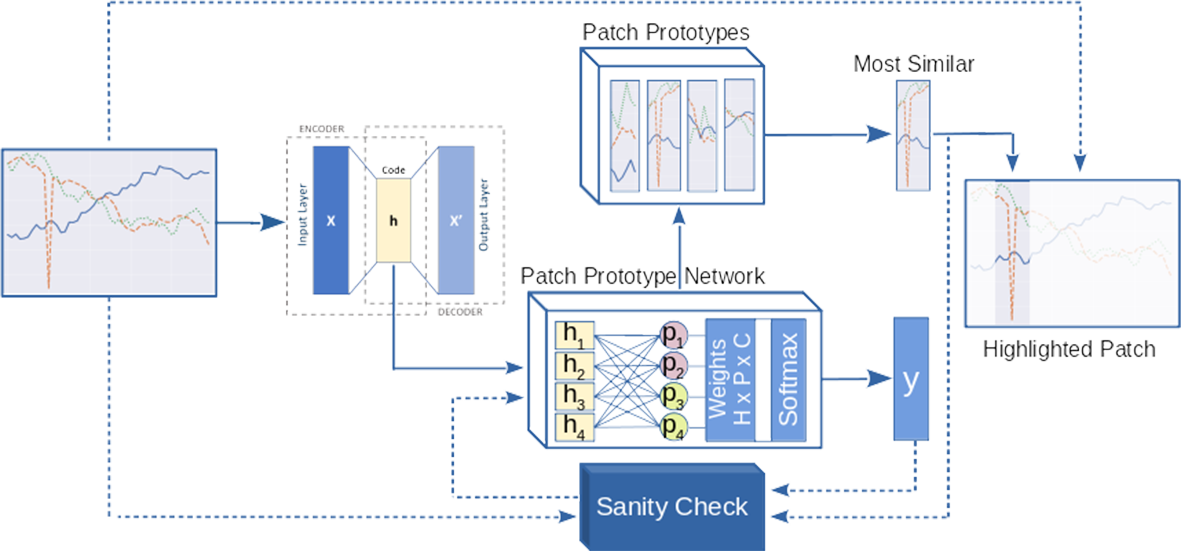}
\caption{\textbf{Inference and testing workflow.} Artificially, computed prototypes are evaluated in a similarity-based manner to suggest class-specific patches.}
\label{fig:network_structure}
\end{figure}

Inspired by the work of Gee et al.~\cite{gee2019explaining}, we combined an autoencoder with a prototype network. The autoencoder consists of several convolutional and max-pooling layers serving as a feature encoding network to provide a latent representation that encodes the relevant features of an input sequence. This representation is fed forward to a custom prototype layer to generate prototypes. Motivated by the work of Chen et al.~\cite{chen2018looks}, we use multiple prototypes to represent a sample rather than a single one for the complete input. Precisely, the prototype layer has randomly initialized variables representing patch prototypes of user-defined size. Larger sizes will result in composed concepts, and smaller sizes result in more basic concepts. On top of the prototype layer, we attached a prototype-weight layer to encourage class-specific prototypes and weight their position within the sample to cover the local importance. Finally, a soft-max classification evaluates the similarity scores produced by the prototype layer multiplied with weights of the prototypes, as shown in Figure~\ref{fig:network_structure}.

\subsection{Mathematical Background}
Our method uses a novel combined loss that captures several aspects enabling the network to produce a meaningful set of patch prototypes based on the losses proposed by~\cite{chen2018looks, gee2019explaining}. For the following equations, let $S_{x}$ be the set of patches corresponding to a sample \textit{x} and the set \textit{P} of prototypes.

\subsubsection{Distances:} We use the $L^{2}$ norm to compute the distance between any two vectors. Furthermore, we compute the minimum distance between a sample and any prototype ($D_{s2p}$) and vice versa ($D_{p2s}$). We denote $D_{p2p}$ as the minimal distance between a prototype and all others and calculate the minimum distance to a prototype of the same class $D_{clst}$ and to the other classes $D_{sep}$ w.r.t. \textit{y}. Therefore $P_y$ denotes the subset of \textit{P} assigned to the class label of \textit{y}. The distances are shown in Equations~\ref{eq:d_s2p} to~\ref{eq:sep}.

\noindent\begin{minipage}{.5\linewidth}
\begin{equation}
D_{s2p}(s) = \min_{p \in P} L^{2}(s, p)
\label{eq:d_s2p}
\end{equation}
\end{minipage}%
\begin{minipage}{.5\linewidth}
\begin{equation}
D_{p2p}(p) = \min_{{p}' \in P} L^{2}(p, {p}')
\label{eq:d_p2p}
\end{equation}
\end{minipage}

\noindent\begin{minipage}{.5\linewidth}
\begin{equation}
D_{clst}(s,y) = \min_{p \in P_{y}} L^{2}(s,p)
\label{eq:clst}
\end{equation}
\end{minipage}%
\begin{minipage}{.5\linewidth}
\begin{equation}
D_{sep}(s,y) = \min_{p \in \{ P \setminus P_{y} \}} D_{s2p}(s,p)
\label{eq:sep}
\end{equation}
\end{minipage}

\subsubsection{Loses:} 
To ensure high-quality prototypes, we introduce our novel patch loss. This loss is a combination of different objectives to achieve good accuracy and an explanation that does not contain duplicates or prototypes that are not class-specific. Our loss combines the following losses:

\begin{itemize}
    \item \textbf{Autoencoder loss:} MSE is used to encourage reconstruction later used for prototype reconstruction.
    
    \item\textbf{Classification loss:} To produce logits for the softmax cross-entropy we multiply the reciprocal of $D_{s2p}$ and the prototype-weight layer.
    
    \item\textbf{$L_{p2s}$ and $L_{s2p}$:} These losses preserve the relation between the input and the prototypes and vice versa as shown in Equation~\ref{eq:p2s_l} and~\ref{eq:s2p_l}.    

    \item\textbf{$L_{div}$:} The diversity among the patch prototypes is computed as shown in Equation~\ref{eq:d_l}.
    
    \item\textbf{$L_{clst}$ and $L_{sep}$:} To encourage the network to learn class-specific prototypes we compute $L_{clst}$ and similarly to $L_{sep}$ but with a negative sign. This penalized prototypes close to samples of the wrong class w.r.t. their assigned class.
\end{itemize}

\noindent\begin{minipage}{.5\linewidth}
\begin{equation}
\small
L_{p2s}(x) = \frac{1}{\left | P \right |} \sum_{p \in P} D_{p2s}(p)
\label{eq:p2s_l}
\end{equation}
\end{minipage}%
\begin{minipage}{.5\linewidth}
\begin{equation}
L_{s2p}(x) = \frac{1}{\left | S_{x} \right |} \sum_{s \in S_{x}} D_{s2p}(s)
\label{eq:s2p_l}
\end{equation}
\end{minipage}

\noindent\begin{minipage}{.5\linewidth}
\begin{equation}
\small
L_{div} = \log (1 + \frac{1}{ \left | P \right |}\sum_{p \in P} D_{p2p}(p))^{-1}
\label{eq:d_l}
\end{equation}
\end{minipage}%
\begin{minipage}{.5\linewidth}
\begin{equation}
L_{clst}(x,y) = \frac{1}{\left | S_{x} \right |} \sum_{s \in S_{x}} D_{clst}(s,y)
\label{eq:clst_l}
\end{equation}
\end{minipage}

Our proposed final loss is a linear combination taking into account previously mentioned aspects and ensures meaningful, diverse, and class-specific patch prototypes shown in Equation~\ref{eq:l}. By default, we set all lambda values except $\lambda_{c}$ to one to find the best compromise between the objectives preserving high accuracy.

\begin{multline}
\small
Patch\_Loss(x,y) = \lambda_{c} H(x,y) + \lambda_{mse} MSE(x,x) + \lambda_{p2s} L_{p2s}(x) \\ 
+ \lambda_{s2p} L_{s2p}(x) + \lambda_{div} L_{div} + \lambda_{clst} L_{clst}(x,y) + \lambda_{sep} L_{sep}(x,y)
\label{eq:l}
\end{multline}

\subsection{Training Process}
The training process of your approach consists of two stages. In the first stage, we fix the weights of the pre-initialized prototype-weight layer to ensure class-specific prototypes. We then train the network until it converges. In the second learning phase, all layers except the prototype-weighting layer are frozen, and the network learns to adjust the prototype weights.  The adjustment corrects the prototype class affiliation using the previously trained latent representation.

\section{Datasets}
We used eight publicly available time-series datasets to emphasize the broad applicability of our approach and examine possible limitations. As a representative set, we used seven different datasets from the UCR Time Series Classification Repository\footnote{http://www.timeseriesclassification.com/} and a point anomaly dataset proposed in~\cite{siddiqui2019tsviz}. These datasets and their parameters are visualized in Table~\ref{tab:accuracy_tradeoff}. Note that the Devices dataset corresponds to the 'Electrical devices' dataset taken from the UCR. To have better coverage of different types, we selected the datasets based on the characteristics concerning the number of classes, channels, and time-steps to cover several conditions and show the prototypes. However, we focus on classification datasets.

\section{Experiments}
In this section, we present our results concerning the performance, applicability, and resource consumption for our proposed approach, highlighting a comparable performance while producing interpretable results.

\subsection{P2ExNet: Instance-based Evaluation}
\begin{figure}[!t]
\centering
\subfloat[Original]{
    \includegraphics[width=0.235\linewidth]{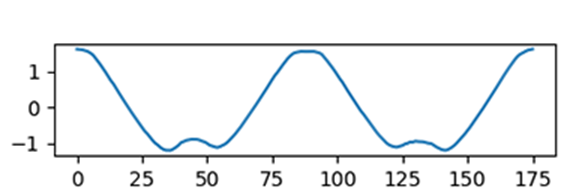}
\label{fig:adiac_13_o}
}
\hfil
\subfloat[Modified]{
    \includegraphics[width=0.235\linewidth]{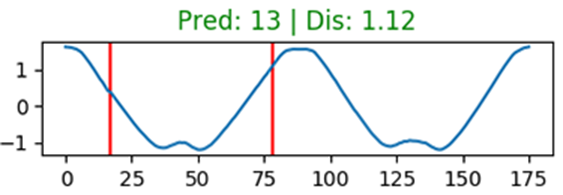}
\label{fig:adiac_13_m}
}
\hfil
\subfloat[Prototypes]{
    \includegraphics[width=0.47\linewidth]{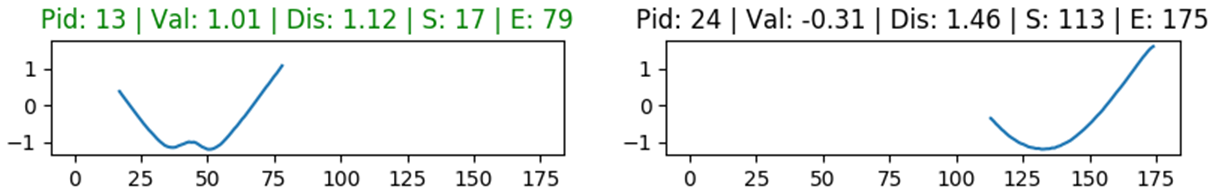}
\label{fig:adiac_13_p}
}
\caption{\textbf{Adiac dataset prototype explanation.} a) shows the original series. b) shows the series with the prototype between the red bars. c) shows two prototypes.}
\label{fig:adiac_13}
\end{figure}

\begin{figure}[!t]
\centering
\subfloat[Time-series]{
\includegraphics[width=0.48\linewidth]{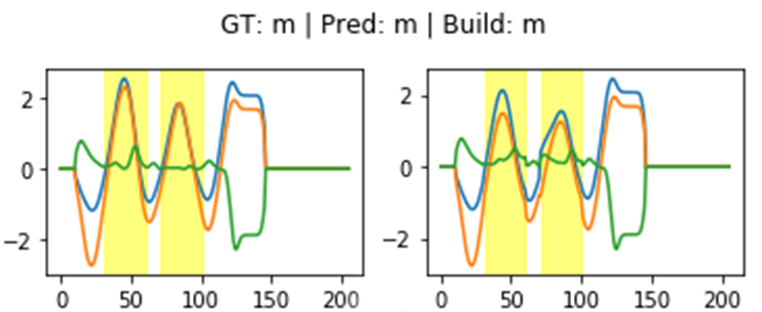}
\label{fig:character_trajectories_m_o}
}
\hfil
\subfloat[Character of the class 'm']{
\includegraphics[width=0.48\linewidth]{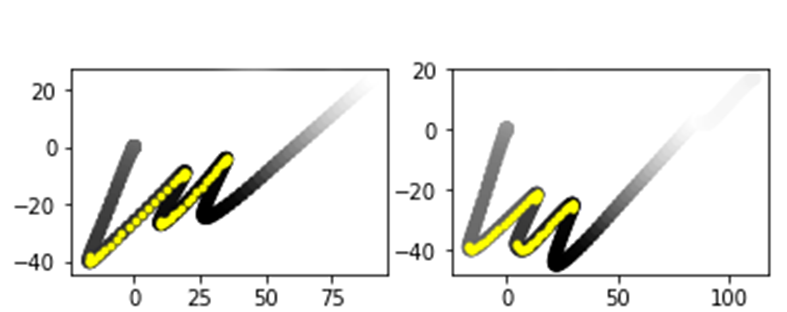}
\label{fig:character_trajectories_m_p}
}
\caption{\textbf{Character dataset prototype explanation.} a) shows the original series and the series with the prototypes. b) shows the character output and the modified character.}
\label{fig:character_trajectories_m}
\end{figure}

\begin{figure}[!t]
\centering
\subfloat[Overall distribution]{
\includegraphics[width=0.39\linewidth]{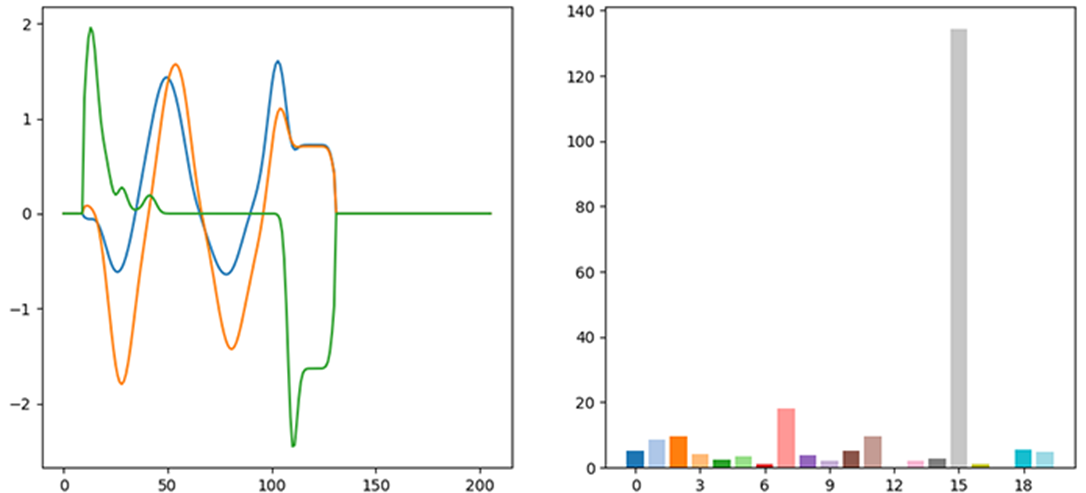}
\label{fig:character_trajectories_dist}
}
\hfil
\subfloat[Patch distribution]{
\includegraphics[width=0.48\linewidth]{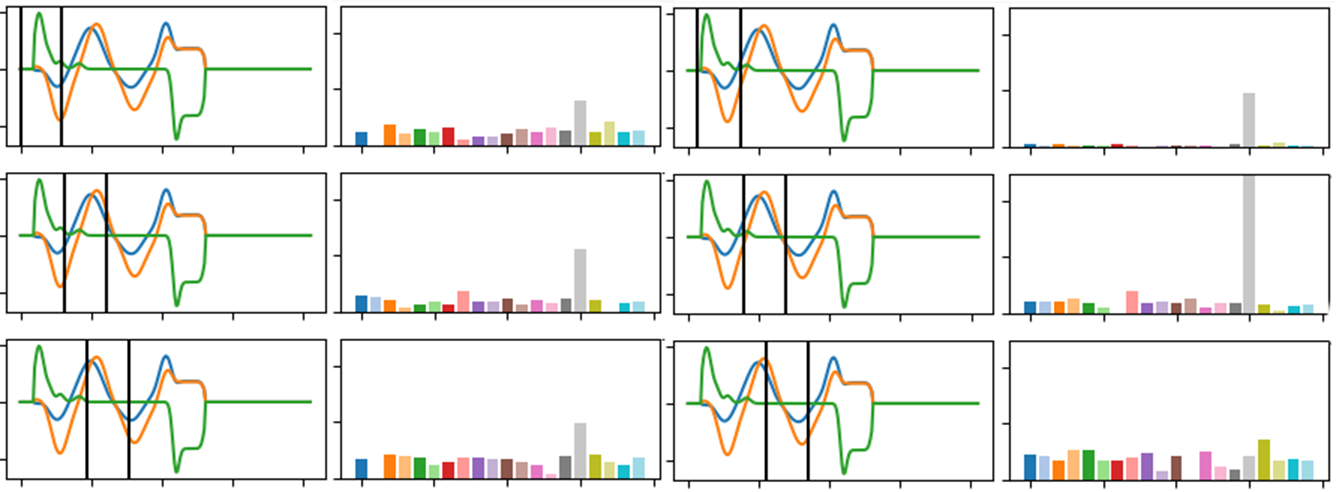}
\label{fig:character_trajectories_dist_detail}
}
\caption{\textbf{Class and prototype distribution.} a) shows the class similarities. b) shows some patches and the corresponding class similarities.}
\label{fig:character_trajectories_distribution}
\end{figure}

\begin{figure}[!t]
\centering
\subfloat[Original]{
    \includegraphics[width=0.23\linewidth]{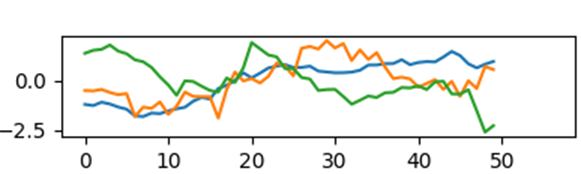}
\label{fig:anomaly_new_1_o}
}
\hfil
\subfloat[Modified]{
    \includegraphics[width=0.23\linewidth]{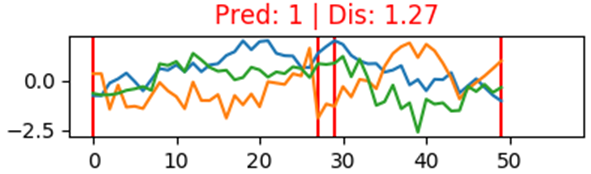}
\label{fig:anomaly_new_1_m}
}
\hfil
\subfloat[Original]{
\includegraphics[width=0.23\linewidth]{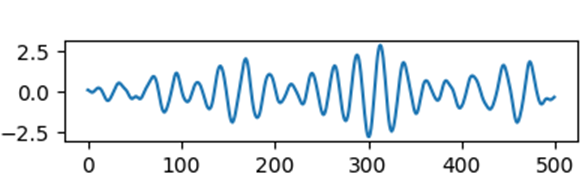}
\label{fig:fordA_0_o}
}
\hfil
\subfloat[Modified]{
\includegraphics[width=0.23\linewidth]{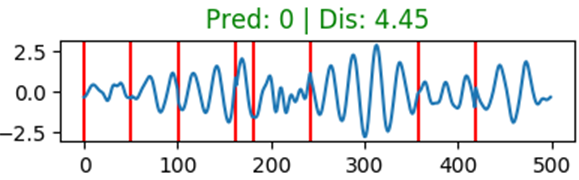}
\label{fig:fordA_0_m}
}
\caption{\textbf{Prototype substitution.} a) and c) show original time-series. b) and d) show the corresponding modified samples and their re-classification.}
\label{fig:substitution}
\end{figure}

The proposed method provides the possibility to identify and highlight the parts of the input that were most relevant for the classification. Besides, it provides prototypes along with a sample containing the prototypes to compare it to the original input. Figure~\ref{fig:adiac_13} shows highlighted regions that were important for the inference on the ADIAC dataset sample. This explanation includes the original sample of the adiac dataset, a modified version, and two prototypes. In the modified version shown in Figure~\ref{fig:adiac_13_m}, we replaced the part between the two red lines with the most important patch prototype to show how close it is to the original part. Figure~\ref{fig:adiac_13_p} shows two prototypes. The value of each prototype denoted as 'Val' highlights its contribution towards the classification result. Similarly, Figure~\ref{fig:character_trajectories_m} shows a sample from the character trajectories dataset and the mapping of the time-series back to the character. The black value highlights the pressure of the pen, and the yellow part shows the mapping of the prototype back to the input space. 
In the case of an incorrect classification, the prototypes have a red caption. Furthermore, in Figure~\ref{fig:character_trajectories_distribution} the class-wise overall and patch-wise distribution provides additional information about similar classes and important patch positions. Especially in Figure~\ref{fig:character_trajectories_dist_detail}, we show that not all patches have the same importance when it comes to the classification. There are sensitive datasets for which the re-classification can change if the original data gets replaced with a prototype. However, for the classification and the explanation, this is not a problem as it can be solved. A proper re-scaling and adjustment can remove the offset between the prototype and the time-series. In Figure~\ref{fig:anomaly_new_1_m} such a jump in the orange signal is shown and leads to an anomaly. However, the classification of the original signal with the network was correct. Furthermore, some datasets are invariant to small offsets shown in Figure~\ref{fig:fordA_0_m}. That is why re-scaling should be done based on the problem task. In case of a point anomaly task, the patches have to align. In a classification task, it is unlikely that the offset of a single point changes the prediction.

\subsection{P2ExNet: Evaluation as a Classifier}
Usually, intrinsic interpretability approaches come with an accuracy drop. In Table~\ref{tab:accuracy_tradeoff} we present the accuracy trade-off highlighting that our structure is on the same level as the non-interpretable counterpart. To create a network similar to ours without the interpretable part, we replaced the prototype layer with a dense layer and a cross-entropy loss, as suggested by Chen et al.~\cite{chen2018looks}. Furthermore, we removed the decoder as there is no need to restrict the latent representation as no reconstruction is required. We conducted this comparison for all eight datasets showing that P2ExNet achieves comparable or better performance in comparison to the non-interpretable variant. Overall the interpretable network has an insignificant performance increase of 0.03\%. Each architecture was superior in four out of the eight datasets. The results show that the accuracy using the interpretable model dropped about 6\% on the anomaly dataset but increased 7\% on the Electric Devices dataset.

\begin{table}[!t]
\caption{\textbf{Accuracy comparison}. A comparison of interpretable and the corresponding non-interpretable counterpart.}
\label{tab:accuracy_tradeoff}
\centering
\begin{tabular}{|c|c|c|c|c|c|}
\hline
\textbf{Dataset} & \textbf{Classes} & \textbf{Length} & \textbf{Channel} & \textbf{CNN} & \textbf{P2ExNet} \\
\hline
Anomaly~\cite{siddiqui2019tsviz} & 2 & 50 & 3 & \textbf{99.79} & 93.79 \\
FordA & 2 & 500 & 1 & 85.44 & \textbf{89.32} \\
Devices & 7 & 96 & 1 & 55.42 & \textbf{62.53} \\
\hline
\hline
Adiac & 37 & 176 & 1 & \textbf{63.54} & 60.15 \\
Crop & 24 & 46 & 1 & 68.27 & \textbf{68.54} \\
\hline
\hline
50words & 13 & 270 & 1 & 76.84 & \textbf{81.98} \\
PenDigits & 10 & 8 & 2 & \textbf{94.29} & 93.95 \\
Character & 20 & 206 & 3 & \textbf{96.53} & 91.78 \\
\hline
\end{tabular}
\end{table}

\subsection{P2ExNet: Sanity Check}

\begin{table}[!t]
\caption{\textbf{Replacement of original patch.} The second column shows how much data was replaced with the suggested prototypes proposed by P2ExNet. The third column shows whether the prediction was the same as with the original time-series or not. The fourth column shows the P2ExNet accuracy for the original sample and the last column for the sample replacing the original patch with the suggested patch. The first row of each dataset corresponds to replacements with the most similar whereas the second row with the most different prototype.}
\label{tab:sanity}
\centering
\begin{tabular}{|c|c|c|c|c|}
\hline
\textbf{Dataset} & \textbf{Data replaced} & \textbf{Equal Pred.} & \textbf{P2ExNet Acc.} & \textbf{P2ExNet mod. Acc.} \\
\hline
\multirow{2}{*}{Anomaly} & 71.99 & 87.43 & \multirow{2}{*}{93.79} & \textbf{91.78} \\
                         & 67.32 & 19.45 &                          & 22.72 \\
\hline
\multirow{2}{*}{FordA} & 51.17 & 99.92 & \multirow{2}{*}{89.32} & \textbf{89.40} \\
                       & 44.95 & 23.09 &                        & 32.69 \\
\hline
\multirow{2}{*}{Devices} & 52.36 & 81.65 & \multirow{2}{*}{62.53} & \textbf{60.52} \\
                         & 65.81 & 49.81 &                          & 39.11 \\
\hline
\hline
\multirow{2}{*}{Adiac} & 35.22 & 85.97 & \multirow{2}{*}{60.15} & \textbf{55.98} \\
                       & 69.90 & 9.11 &                            & 14.84 \\
\hline
\multirow{2}{*}{Crop} & 50.50 & 94.08 & \multirow{2}{*}{68.54} & \textbf{66.94} \\
                     & 81.12 & 22.01 &                          & 23.28 \\
\hline
\hline
\multirow{2}{*}{50words} & 36.43 & 93.01 & \multirow{2}{*}{81.98} & \textbf{77.20} \\
                         & 52.88 & 62.50 &                          & 56.98 \\
\hline
\multirow{2}{*}{PenDigits} & 69.47 & 99.31 & \multirow{2}{*}{93.95} & \textbf{93.54} \\    
                           & 68.65 & 8.83 &                        & 11.0 \\
\hline
\multirow{2}{*}{Character} & 18.15 & 92.93 & \multirow{2}{*}{91.78} & \textbf{85.30} \\
                           & 52.90 & 31.71 &                        & 32.87 \\
\hline
\end{tabular}
\end{table}

\begin{table}[!t]
\caption{\textbf{Closeness of prototypes.} The difference between representative and generated latent patch prototypes for P2ExNet with and without the use of the decoder are shown.}
\label{tab:prototype_stats}
\centering
\begin{tabular}{|c|c|c|c|}
\hline
\textbf{Dataset} & \textbf{P2ExNet with decoder} & \textbf{P2ExNet without decoder} & \textbf{Improvement} \\
\hline
Anomaly & 0.6393 & \textbf{0.4929} & -22.9\% \\
\hline                 
FordA & \textbf{0.7018} & 1.0315 & 47.0\% \\
\hline
Devices & 0.4135 & \textbf{0.3399} & -17.8\% \\
\hline
\hline
Adiac  & 0.538 & \textbf{0.4993} & -6.2\% \\
\hline
Crop    & \textbf{0.442} & 0.4815 & 8.9\% \\    
\hline
\hline
50words & \textbf{0.0413} & 0.2086 & 505.1\% \\
\hline
PenDigits & \textbf{0.5123} & 0.5622 & 9.7\% \\
\hline
Character & \textbf{0.0099} & 0.5887 & 5946.5\% \\
\hline
\end{tabular}
\end{table}

\begin{figure}[!t]
\centering
\subfloat[Crop dataset]{
\includegraphics[width=0.44\linewidth]{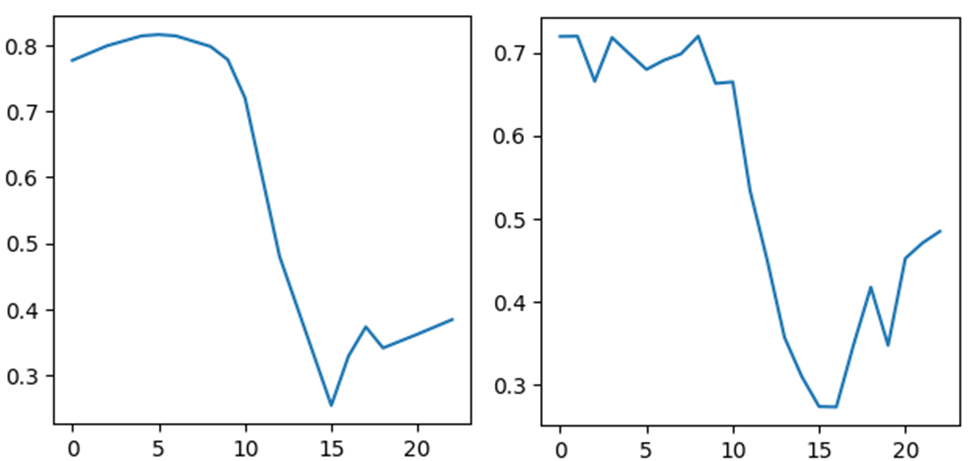}
\label{fig:crop_p0}
}
\hfil
\subfloat[Character trajectories dataset]{
\includegraphics[width=0.44\linewidth]{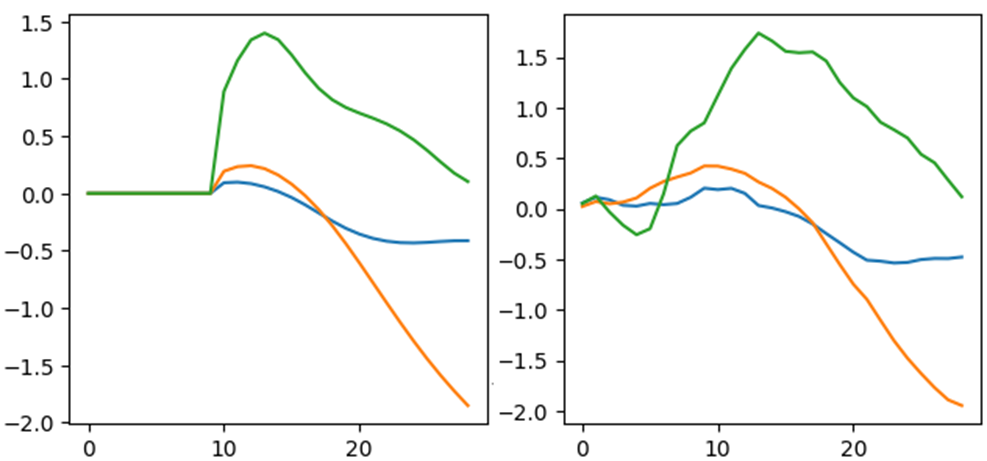}
\label{fig:character_p2}
}
\caption{\textbf{Prototype comparison.} This figure shows the representative patch based on the distance to the latent prototype and the reconstruction of the latent representation.}
\label{fig:prototype_comparison}
\end{figure}

\begin{figure}[!t]
\centering
\includegraphics[width=0.88\linewidth]{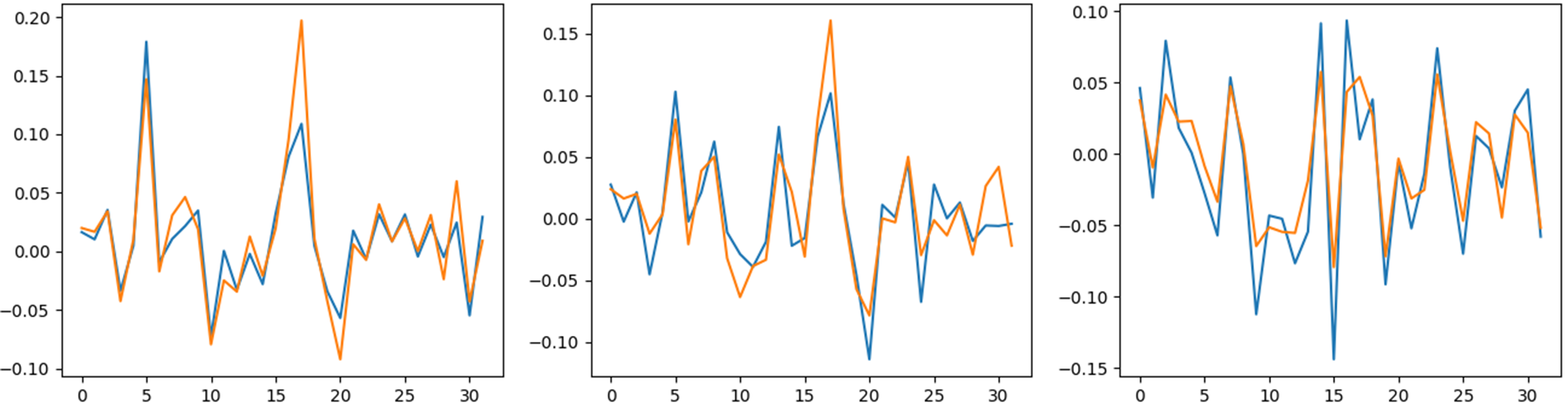}
\caption{\textbf{Latent space difference.} The difference between the prototype (orange) and the real sample (blue) in the latent space for each channel are shown.}
\label{fig:character_p2_latent}
\end{figure}

To prove the class-specific and meaningful behavior of the prototypes, we replaced the original time-series once with the most positive and once with the most negative influencing prototypes. In Table~\ref{tab:sanity} we show that the replacement with the most confident prototypes corresponding to the predicted class achieved results close to the default accuracy, whereas the best fit prototype of a different class dramatically decreased the performance as the prediction switched. These results show that our prototypes are class-specific. However, we conducted the second sanity check to investigate the need for the decoder to produce latent representations that are close to the representative prototypes. In Table~\ref{tab:prototype_stats} we show that for the character trajectories, 50words, and the FordA dataset there is a significant difference if the decoder gets excluded. Also, we compared the representative and decoded prototypes and visualized two prototypes in Figure~\ref{fig:prototype_comparison} highlighting the small difference between the selected representative sample (left) and the decoded one (right). We further provide the latent representation of the character trajectory prototype in Figure~\ref{fig:character_p2_latent}. Each plot represents one of the three channels and the blue color encodes the part of the selected sample whereas the orange color decodes the latent representation of the prototype. It is clearly visible that both latent representations share the same pattern and therefore result in a similar decoded prepresentation as shown in Figure~\ref{fig:character_p2}.

\subsection{Comparison with Existing Prototype-based Approaches}

\begin{figure}[!t]
\centering
\subfloat[Original]{
    \includegraphics[width=0.2\linewidth]{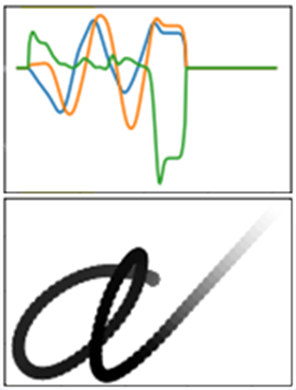}
\label{fig:compare_methods_org}
}
\hfil
\subfloat[Gee et al.~\cite{gee2019explaining}]{
    \includegraphics[width=0.2\linewidth]{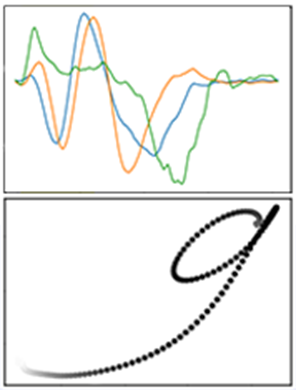}
\label{fig:compare_methods_g}
}
\hfil
\subfloat[Chen et al.~\cite{chen2018looks}]{
    \includegraphics[width=0.2\linewidth]{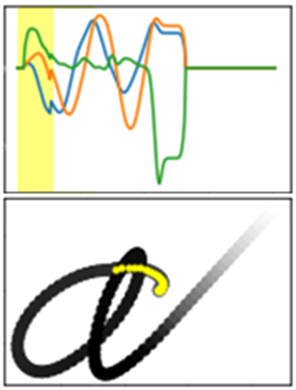}
\label{fig:compare_methods_c}
}
\hfil
\subfloat[P2ExNet]{
    \includegraphics[width=0.2\linewidth]{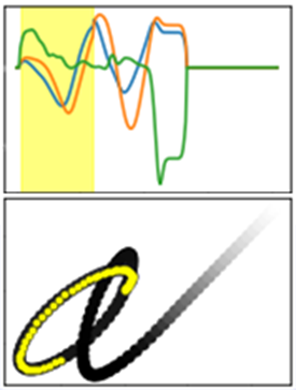}
\label{fig:compare_methods_o}
}
\caption{\textbf{P2ExNet approaches.} Different explanations of the character 'a'.}
\label{fig:compare_methods}
\end{figure}

We compared the proposed method against existing work~\cite{chen2018looks} and~\cite{gee2019explaining}. Precisely, we highlight the explanations and additional outputs. In Figure~\ref{fig:compare_methods} we show the explanation of each approach for a character 'a' sample. While~\cite{gee2019explaining} explains the class with a prototype providing a single prototype capturing the complete sample,~\cite{chen2018looks} is based on parts of the input leading to a more detailed explanation. This method searches a patch for a region in the input image. Precisely, this means additional position information is available. Lastly, our proposed method provides the same information about the location but offers re-scaling as well as an implicit comparison to other prototypes and a class distribution for the complete sample and the patches, as shown in Figure~\ref{fig:character_trajectories_dist_detail}. Furthermore, our prototypes are class-specific and invertible. It is possible to decode them for a comparison with the representatives.

\section{Conclusion}
Summarizing our results, we came up with novel network architecture, along with a loss and training procedure aligned to produce interpretable results and an inference process similar to the human reasoning without a significant drop in performance. Further, we proved that the proposed method works for several time-series classification tasks and when excluding the class-specific prototype assignment, our approach is suitable to produce prototypes for regression and forecast tasks. Besides, we compared the proposed method with existing prototype-based methods concerning their interpretable output and time consumption, finding ours superior in both aspects.

\section*{Acknowledgements}
This work was supported by the BMBF projects DeFuseNN (Grant 01IW17002) and the ExplAINN (BMBF Grant 01IS19074). We thank all members of the Deep Learning Competence Center at the DFKI for their comments and support.

\end{document}